\documentclass[runningheads]{llncs}
\usepackage{lmodern}

\usepackage[mobile]{eccv}

\usepackage{eccvabbrv}

\usepackage[accsupp]{axessibility}  
\usepackage{worldflags}

\usepackage[dvipsnames]{xcolor}
\usepackage{booktabs}
\usepackage{graphicx}
\usepackage{amsmath}
\usepackage{amssymb}
\usepackage{algorithm}
\usepackage{algpseudocode}
\usepackage{subcaption}
\usepackage{arydshln}

\usepackage{epigraph}

\usepackage[font={small}]{caption}

\usepackage{microtype}
\usepackage{duckuments} 

\newcommand{\typicality}{{\mathbf{T}}}
\newcommand{\tbar}{\bar{{\typicality}}}

\def\eqref#1{Eq.~(\ref{#1})}

\usepackage{xstring}

\newcommand{\processstring}[1]{%
  \StrSubstitute{#1}{\_}{\space}[\temp]
  \StrLeft{\temp}{1}[\first]
  \StrGobbleLeft{\temp}{1}[\rest]
  \expandafter\MakeUppercase\first\rest
}
\def\basefigem{-1.1em}

\def\trange{0.1-0.7}

\def\carcla{1920}
\def\carclb{1930}

\def\carcld{1980}
\def\topkcarsa{1930}
\def\topkcarsb{1990}

\def\fttclb{1920}
\def\fttclc{1940}
\def\fttcld{1950}
\def\topkftta{1920}
\def\topkfttb{1970}

\def\geocla{United States}
\def\geoclb{Russia}
\def\geoclc{Brazil}
\def\geocld{France}
\def\geocle{Japan}
\def\geoclf{Thailand}
\def\topkgeoa{Thailand}
\def\topkgeob{Russia}

\def\placesa{basketball\_court\_indoor}
\def\placesb{chemistry\_lab}
\def\placesc{construction\_site}
\def\placesd{beer\_garden}
\def\placese{veterinarians\_office}
\def\placesf{stage\_outdoor}

\usepackage{hyperref}

\usepackage{orcidlink}

\begin{document}

\title{Diffusion Models as Data Mining Tools} 

\titlerunning{Diffusion Models as Data Mining Tools}

\author{Ioannis Siglidis\inst{1}\orcidlink{0009-0002-2278-5825} \and
Aleksander Holynski\inst{2}\orcidlink{0009-0008-6915-0126} \and
Alexei A. Efros\inst{2}\orcidlink{0000-0001-5720-8070} \and \\
Mathieu Aubry\inst{1}\orcidlink{0000-0002-3804-0193} \and 
Shiry Ginosar\inst{2}\orcidlink{0000-0002-7362-1401}
}
\authorrunning{I.~Siglidis \etal}

\institute{LIGM, Ecole des Ponts, Univ Gustave Eiffel, CNRS, Marne-la-Valle, France \and University of California, Berkeley\\
\url{https://diff-mining.github.io/}}

\maketitle

\begin{figure}
    \centering
    \includegraphics[width=1.0\textwidth]{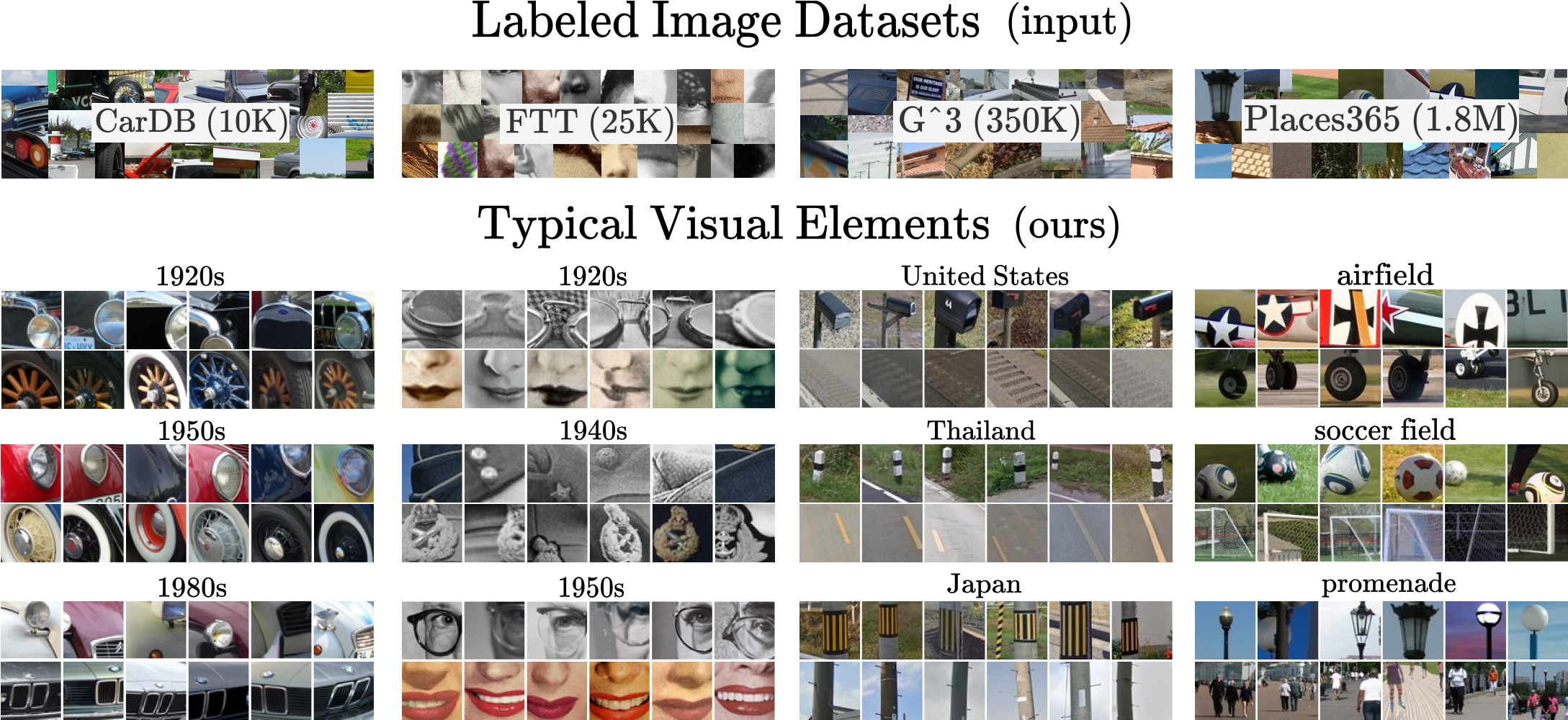}
    \caption{\textbf{Mining typical visual elements with diffusion models.}
    We demonstrate how to use diffusion models to mine visual data through a simple pixel-based score and a standard clustering approach. We present high-quality mining results for a diverse range of datasets (from left to right: 10,130 photographs of cars tagged with a creation year between 1920-1999~\cite{cardb}, 24,874 portraits from the 19th to the 21st century~\cite{ftt}, 344,224 Street View images tagged with country names~\cite{g3}, and {1,803,460 images of scenes images associated with descriptive names~\cite{places365}}). Our results highlight both expected elements and more unforeseen ones.
    }
    \label{fig:teaser}
\vspace{-2em}
\end{figure}

\begin{abstract}
    This paper demonstrates how to use generative models trained for image synthesis as tools for visual data mining. Our insight is that since contemporary generative models learn an accurate representation of their training data, we can use them to summarize the data by mining for visual patterns. Concretely, we show that after finetuning conditional diffusion models to synthesize images from a specific dataset, we can use these models to define a typicality measure on that dataset. This measure assesses how typical visual elements are for different data labels, such as geographic location, time stamps, semantic labels, or even the presence of a disease. This analysis-by-synthesis approach to data mining has two key advantages. First, it scales much better than traditional correspondence-based approaches since it does not require explicitly comparing all pairs of visual elements. Second, while most previous works on visual data mining focus on a single dataset, our approach works on diverse datasets in terms of content and scale, including a historical car dataset, a historical face dataset, a large worldwide street-view dataset, and an even larger scene dataset. Furthermore, our approach allows for translating visual elements across class labels and analyzing consistent changes. Project page: \url{https://diff-mining.github.io/}.
    \keywords{Visual Data Mining \and Diffusion Models}
\end{abstract}

\section{Introduction}
\label{sec:intro}
Visual data mining aims to discover patterns within large visual corpora such as collections of street view panoramas~\cite{paris2015,linking2015iccp}, historical images {of faces}~\cite{ginosar2017yearbooks,ftt} or photographs of {cars}~\cite{cardb, dalens2019bilinear}. This paper proposes a novel idea: to turn generative models trained for image synthesis into a scalable method for visually mining image datasets. Generative models digest massive amounts of data, which they implicitly store in their weights.
Our central insight is that we can use this learned summary of the visual input to identify the most typical image regions. This unconventional use of a diffusion model for studying its training data demonstrates that generative models are potent tools beyond synthesis---for data mining, summary, and understanding.

Our target task, mining for informative visual patterns, is challenging. Unlike text, where words act as discrete tokens that we can directly compare, the visual world seldom contains exactly repeating elements. Even common simple visual elements, such as windows, can have different colors and different numbers of panes; they may be seen from various viewpoints, and they may be located at multiple positions as part of different facades. The standard approach to visual data mining~\cite{paris2015,cardb,shen2021learning} involves learning data-specific similarities with relevant invariances (\eg, such that different-looking windows will be similar) and using them to search for discriminative patterns. However, these techniques are not easily scalable since one must apply them across all pairs of visual elements within all pairs of images in the dataset. 
The similarity graph between visual elements scales quadratically with the size of the dataset. 
In contrast, our proposed analysis-by-synthesis approach does not require pairwise comparisons between different visual elements and thus scales to very large datasets. 

The approach we propose takes as input a dataset with image-level tags, such as time~\cite{cardb,ftt}, geography~\cite{g3}, or scene labels~\cite{places365}. Our goal is to provide a visual summary of the elements typical of the different tags, such as the common elements that enable us to determine the location of a streetview panorama. To arrive at this summary, we first finetune a conditional diffusion model on the target dataset. We then use the finetuned model to define a pixel-wise typicality measure by assessing the degree to which the label conditioning impacts the model's reconstruction of an image. We mine visual elements by aggregating typicality on patches, selecting the most typical ones, and clustering them using {features extracted from the finetuned model}~\cite{dift}. As visualized in Fig.~\ref{fig:teaser}, this leads to clusters of typical visual elements that summarize the most characteristic patterns associated with the tags available in the input dataset. For example, our face results highlight iconic elements, such as aviator glasses in the 1920s and {military hats} in the {1940s}, and more subtle details, such as period-typical glasses or make-up. Interestingly, our results on street-view data highlight details that are similar to the ones presented in geographical understanding websites~\cite{geodummy,geohints,plonkit}, popularized through the GeoGuessr game~\cite{geoguessr}, such as typical parts of {utility poles, bollards, or architecture}. To our knowledge, no existing visual mining method has demonstrated such high-quality results on diverse datasets.
\section{Related Work}

\medskip
\noindent
\textbf{Visual data mining.}
Visual data mining turned the manual and subjective process of comparing photographs (\eg,~\cite{kotchemidova2005we}) into algorithmic methods for summarizing image data, such as architectural details~\cite{paris2015,linking2015iccp}, fashion~\cite{ginosar2017yearbooks,StreetStyle2017,ftt}, industrial design~\cite{jae2013style}, and art~\cite{shen2019discovering,shen2021large,kaoua2021imagecollation} by locating visual patterns. This has mainly been achieved using discriminative techniques such as clustering or contrastive learning. For example, ~\cite{cardb} demonstrated how correspondence based mining across time can be achieved in a dataset of objects of similar parts, namely cars, and~\cite{paris2015} showed that geographically representative image elements can be automatically discovered from Google Street View imagery in a discriminative manner. 
However, such traditional data mining approaches do not scale to large modern datasets. Indeed, they require pairwise comparisons between all the visual elements of each image to the entire dataset in order to locate nearest neighbors and establish clusters. Notably, the discriminative clustering algorithm of~\cite{paris2015} requires training a separate linear SVM detector for each visual element- a computationally prohibitive approach when considering multiple possible visual elements for the purposes of analysis. In contrast, our approach is scalable to very large datasets.
Closer to our work, generative model have been trained to analyze the evolution of faces~\cite{ftt} and cars~\cite{dalens2019bilinear} across time. However, these two works essentially perform image translation, and do not enable actual mining of typical elements in the datasets.

\medskip
\noindent
\textbf{Diffusion models.}
Diffusion models have gained popularity in recent years due to their stability in training and effectiveness in modeling complex multimodal distributions~\cite{sohl2015deep, ho2020denoising, dhariwal2021diffusion, ho2022cascaded,  karras2022elucidating}. These models are capable of generating high-quality imagery conditioned on input signals beyond categorical labels, like text~\cite{ldm, imagen, dalle2}, and can further incorporate additional modalities~\cite{controlnet, gligen}.
In addition to generating images from scratch, diffusion models have been used {extensively for} instruction-driven image-to-image translation~\cite{sdedit, prompt-to-prompt, nullinversion, tumanyan2023plug, instructpix2pix}.
It has also been shown that pre-trained text-to-image diffusion models encode strong priors for natural scenes, allowing their internal features to be used for secondary tasks~\cite{odise, hyperfeatures, dift}. They can easily be adapted for new tasks or to new data distributions through minimal finetuning~\cite{instructpix2pix, controlnet, dreambooth}.

Beyond mere image synthesis, generative image models, and in particular diffusion models, have been used as data augmentation engines.
While most machine learning approaches treat the data as fixed and improve the learning algorithm, works such as~\cite{jahanian2021generative,chai2021ensembling,azizi2023synthetic} fix the learning algorithm and augment the training data, using generative models to synthesize large amounts of synthetic data. 
In contrast, we present a new way to use generative models, aiming to gain insights about their training data. 

\section{Data Mining via Diffusion Models}

Our approach turns generative models into data mining tools. It relies on finetuning a conditional stable diffusion model trained for image synthesis, {using it to extract a summary of the visual world}. We start by reviewing diffusion models and the techniques we leverage in section~\ref{Preliminary}. In section~\ref{typicallity}, we introduce our measure of typicality, which allows us to measure how the class label conditioning affects the synthesis of an image by the diffusion model. In section~\ref{sec:method-mining}, we describe how we aggregate typicality on patches to mine typical visual elements and cluster them to summarize the training data.

\subsection{Preliminary}
\label{Preliminary}

\medskip
\noindent
\textbf{Diffusion models.}
Diffusion models are generative models trained to transform random noise $\epsilon \sim N(0, 1)$ into a target distribution $\mathcal{X}$. The denoising process is iterative, indexed by a step index $t$. A diffusion model $\epsilon_\theta(z,t)$, with parameters $\theta$, takes as input an image $z$ to be denoised at the fractional timestep $t$.

During training, a training sample $x$ is artificially noised at a strength associated to a uniformly sampled step $t$, by mixing the image with a randomly sampled Gaussian noise image $\epsilon$ in what is known as the \emph{forward process}:
\begin{align}
\text{noise}(x, \epsilon, t) = \sqrt{\bar{a}_{t}}x + (1-\sqrt{\bar{a}_{t}})\epsilon,
\end{align}
where $\sqrt{\bar{a}_t}$ defines the noise mixing coefficient which varies over the denoising process.
The learnable denoising model $\epsilon_\theta$ takes as input both a noised image and the corresponding noising step and is trained to predict the noise image (or equivalently the denoised image) using a loss: 

\begin{align}
L_t(x, \epsilon) = ||\epsilon_\theta(\text{noise}(x, \epsilon, t),\; t) - \epsilon||^2.
\label{eq:Lt}
\end{align}

At test time, the target distribution can be sampled by transforming the noise distribution through an iterative denoising process~\cite{ho2020denoising,song2020denoising}, in which a randomly sampled image of noise $z_T$ is gradually denoised according to the model's predictions. 

\medskip
\noindent
\textbf{Conditional Diffusion Models.} 
Diffusion models can be extended to take a conditioning $c$ associated with the image content as an additional input. This leads to a model $\epsilon_\theta(z,t,c)$ that depends on the noisy image $z$, the time step $t$, the conditioning $c$, and a loss $L_t(x,\epsilon,c)$. In our case, $c$ will correspond to the CLIP~\cite{clip} text features of the class label associated with the image. 

\medskip
\noindent
\textbf{Latent diffusion models.} We use a variant of diffusion models known as a \emph{latent} diffusion model (LDM)~\cite{ldm}. Instead of modeling the source data distribution, LDMs model its distribution in the latent space of a variational autoencoder~\cite{kingma}. Working in the latent space reduces the complexity of the data distribution. It thus significantly reduces both the number of parameters of the diffusion model and the amount of training samples necessary to learn a good model.

\subsection{Typicality}
\label{typicallity}

We design our measure of typicality based on the following intuition: a visual element is {typical} of a conditioning class label (\eg, country name or date) if the diffusion model is better at denoising the input image in the presence of the label than in its absence. We, therefore, design typicality as a ranking measure across pixels between the ground-truth conditioning $c$ and the null conditioning $\varnothing$. We define the \textit{typicality} of an image $x$ given the class label conditioning $c$ as:
\begin{equation}
\typicality (x|c) =  \mathbb{E}_{\epsilon,t}[L_t(x, \epsilon, \varnothing) - L_t(x, \epsilon, c)],
\end{equation}
where $t$ is sampled uniformly from $[0, 1]$, and $\epsilon$ is sampled according to the noise distribution $N(0,1)$. This typicality measure enables us to sort visual elements from a specific class by how typical they are of that class (see supplementary material for additional formal motivation of this measure). Our typicality measure is related to the image-level classification approach of Li \etal~\cite{deepak}, but it is built for pixel-based analysis and data mining. {Unlike Li \etal, we find that reducing the sampled range of $t$ to $[0.1, 0.7]$ improves the quality of our results, as the tails can contribute uninformative yet typical samples, as we show in the supplementary material.}

\subsection{Mining for Typical Visual Elements}
\label{sec:method-mining}

\medskip
\noindent
\textbf{Conditioning and finetuning.}
To mine typical visual elements for a given class, we use the text class label conditioning $c$ in the form of its CLIP text features~\cite{clip}. We convert the tags associated with the datasets to text using the embeddings of the following sentences: “A car/portrait from the \{decade\}s.” for faces and cars (“A car/portrait.” for the {null} conditioning $\varnothing$), “A Google streetview image of \{country\}.” for streetview data (“A Google streetview image.” for the {null} conditioning $\varnothing$), {and “An image of \{scene\}.” for images of the Places dataset~\cite{places365} (empty string for the {null} conditioning $\varnothing$)}. We finetune a latent diffusion model~\cite{ldm} on the target dataset by optimizing the reconstruction loss (Equation~\ref{eq:Lt}) given the conditioning. We use Stable Diffusion V1.5~\cite{ldm} as a base model in all our experiments.

\medskip
\noindent
\textbf{Patch-based analysis.} To find condition-specific visual elements, we compute our typicality scores over patches of images by averaging typicality in the area of a patch\footnote{Since we use latent diffusion, the loss for arbitrary patches requires upsampling the feature maps to the original image resolution.}. To identify the set of most typical visual elements for a dataset we pick the 5 most typical non-overlapping patches in each image according to the patch typicality, and select the 1000 most typical patches over all the dataset.

\medskip
\noindent
\textbf{Clustering visual elements.}
We cluster the most typical patches using k-means~\cite{kmeans} with 32 clusters. {To cluster elements, we embed them with DIFT~\cite{dift} features, computed at timestep $t=0.161$ using our finetuned models.
For visualization, we rank clusters by the median typicality of their elements in decreasing order and the elements within a cluster by the distance to the centroid in increasing order.

\section{Experiments}

\label{sec:exp}
{We showcase the effectiveness of our approach in summarizing visual data for a wide variety of datasets. First, in Section~\ref{ss:dataset}, we introduce the datasets used in our experiments. Second, in section~\ref{ss:typicality-eval}, we evaluate the ranking given by our typicality measure. Third, in Section~\ref{sec:exp-clusters}, we discuss our main result, the mined visual summaries of the analyzed datasets, and compare with Doersch \etal~\cite{paris2015}. Finally, we discuss the limitations of our approach in Section~\ref{ss:lim}.}

\subsection{Datasets}
\label{ss:dataset}

We experiment with four diverse datasets. CarDB~\cite{cardb} and FTT~\cite{ftt} have already been used for visual mining and include a few tens of thousands of images. G\string^3~\cite{g3} and Places~\cite{places365} are much larger with 344K and 1.8M images respectively, and, to our knowledge, have never been used for visual mining.

\medskip
\noindent
\textbf{Cars.} The CarDB dataset \cite{cardb} contains 10,130 photos of cars from 1920 to 1999, collected from \url{cardatabase.net}. They are labeled with creation years, which we bin into decades for our analysis. This dataset contains cars seen from various viewpoints and in diverse environments. As a result, extracting time-informative elements is challenging. We rescale all images to a height of 256 pixels while preserving their original aspect ratio.

\medskip
\noindent
\textbf{Faces.} The Faces Through Time (FTT) Dataset \cite{ftt} contains 24,874  images of notable people from the 19th to 21st century, with roughly 1,900 images per decade, sourced from Wikimedia Commons. All photos are of size 256x256 pixels.

\medskip
\noindent
\textbf{Geo.} The G\string^3~\cite{g3} dataset contains images obtained from crops of street-view panoramas, diversely sampled worldwide, of which we selected 344,224 images, which we rescaled to 512x756 pixels. This dataset is challenging because of the small details that characterize a scene's appearance and scale.  We focus on the 8 countries with the largest number of panoramas (United States, Japan, France, Italy, United Kingdom, Brazil, Russia, and Thailand) and two countries with fewer images (Nigeria and India). We finetune the network using all images from these countries, but we only mine a random subset of 1000 images.

\medskip
\noindent
\textbf{Places.} The high-resolution version of the Places dataset~\cite{places365} contains 1,803,460 million images from 365 place categories associated with their labels, with a minimum dimension of 512 pixels. For mining we only use the validation dataset, which contains 100 images per scene category.

\subsection{Typicality Measure Evaluation}
\label{ss:typicality-eval}
\label{ss:fine-tuning}
\def\topklen{0.23}
\def\topklenb{0.23}

\begin{figure}[t]
    \centering
    \hspace{-4pt}

    \begin{subfigure}{\topklenb\textwidth}
        \centering
        {\scriptsize\;\;\;\;\; 1930s}
        \makebox[7pt]{\hspace{-15pt}\raisebox{3pt}{$\typicality$}}\includegraphics[width=\textwidth,trim={0 0 52px 0},clip]{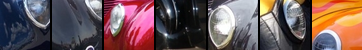}
        \makebox[7pt]{\hspace{-18pt}\raisebox{3pt}{Rand.}}\includegraphics[width=\textwidth,trim={0 0 52px 0},clip]{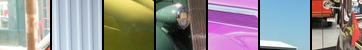}
        \makebox[7pt]{\hspace{-23pt}\raisebox{3pt}{$-\typicality$}}\includegraphics[width=\textwidth,trim={0 0 52px 0},clip]{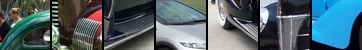}
        \vspace{1pt}
        {\scriptsize\;\;\;\;\; 1990s}
        \makebox[7pt]{\hspace{-15pt}\raisebox{3pt}{$\typicality$}}\includegraphics[width=\textwidth,trim={0 0 52px 0},clip]{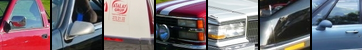}
        \makebox[7pt]{\hspace{-18pt}\raisebox{3pt}{Rand.}}\includegraphics[width=\textwidth,trim={0 0 52px 0},clip]{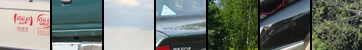}
        \makebox[7pt]{\hspace{-23pt}\raisebox{3pt}{$-\typicality$}}\includegraphics[width=\textwidth,trim={0 0 52px 0},clip]{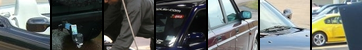}

        \label{topk:cars}
        \caption{\textbf{CarDB} \cite{cardb}}
    \end{subfigure}
    \hspace{4.5pt}
    \begin{subfigure}{\topklenb\textwidth}
        \centering
        {\scriptsize 1920s}
        \includegraphics[width=\textwidth,trim={0 0 52px 0},clip]{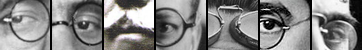}
        \includegraphics[width=\textwidth,trim={0 0 52px 0},clip]{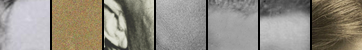}
        \includegraphics[width=\textwidth,trim={0 0 52px 0},clip]{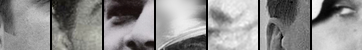}
        {\scriptsize 1970s}
        \vspace{1pt}
        \includegraphics[width=\textwidth,trim={0 0 52px 0},clip]{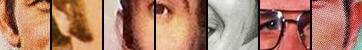}
        \includegraphics[width=\textwidth,trim={0 0 52px 0},clip]{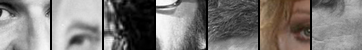}
        \includegraphics[width=\textwidth,trim={0 0 52px 0},clip]{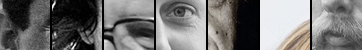}
        
        \label{topk:ftt}
        \caption{\textbf{FTT} \cite{ftt}}
    \end{subfigure}
    \begin{subfigure}{\topklen\textwidth}
        \centering

        {\scriptsize Thailand}
        \includegraphics[width=\textwidth,trim={0 0 66px 0},clip]{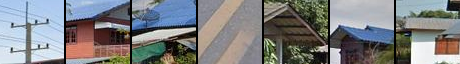}
        \includegraphics[width=\textwidth,trim={0 0 66px 0},clip]{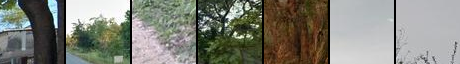}
        \includegraphics[width=\textwidth,trim={0 0 66px 0},clip]{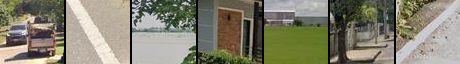}
        \vspace{0pt}
        {\scriptsize Russia}
        \includegraphics[width=\textwidth,trim={0 0 66px 0},clip]{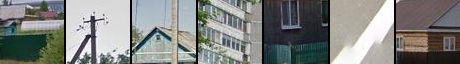}
        \includegraphics[width=\textwidth,trim={0 0 66px 0},clip]{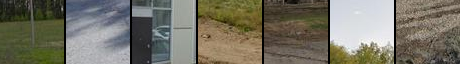}
        \includegraphics[width=\textwidth,trim={0 0 66px 0},clip]{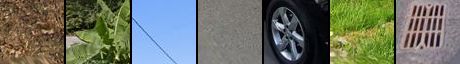}

        \label{topk:geo}
        \caption{\textbf{G}\string^\textbf{3} \cite{g3}}
    \end{subfigure}
    \begin{subfigure}{\topklen\textwidth}
        \centering
        {\scriptsize Soccer Field}
        \includegraphics[width=\textwidth,trim={0 0 66px 0},clip]{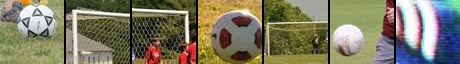}
        \includegraphics[width=\textwidth,trim={0 0 66px 0},clip]{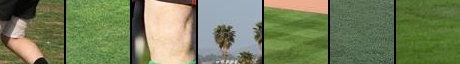}
        \includegraphics[width=\textwidth,trim={0 0 66px 0},clip]{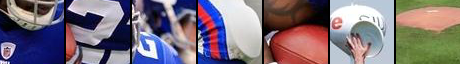}
        \vspace{0pt}
        {\scriptsize Laundromat}
        \includegraphics[width=\textwidth,trim={0 0 66px 0},clip]{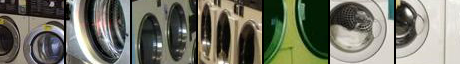}
        \includegraphics[width=\textwidth,trim={0 0 66px 0},clip]{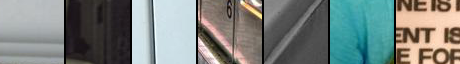}
        \includegraphics[width=\textwidth,trim={0 0 66px 0},clip]{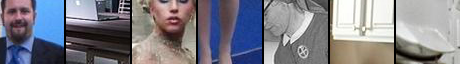}

        \label{topk:places}
        \caption{{\textbf{Places}} \cite{places365}}
    \end{subfigure}
    \caption{\textbf{Typical elements are informative of the conditioning label.} We visualize the top-6 patches ranked according to typicality~($\typicality$) with respect to the conditioning class label, negative typicality ($-\typicality$), and randomly (Rand.). The two rows correspond to different classes from each of the four datasets.
}
    \label{fig:topk}
    \vspace{\basefigem}
\end{figure}


\medskip
\noindent
\textbf{Typicality score for patches.} Fig.~\ref{fig:topk} shows the most and least typical patches according to our typicality measure and random patches for the four datasets. We note that the most typical patches are unique to each class and more discriminative than random patches, while the least typical patches are uninformative of the conditioning label. 

\begin{figure}[!t]
    \begin{subfigure}{0.26\linewidth}
        \centering
       \includegraphics[width=\textwidth, height=16em]{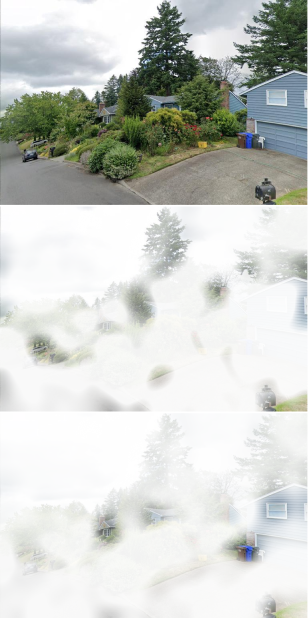}
        \caption{Typicallity}
        \label{fig:finetuning-qualitative:a}
    \end{subfigure}
    \hfill
    \begin{subfigure}{0.42\linewidth}
        \includegraphics[width=1.0\textwidth, height=16em]{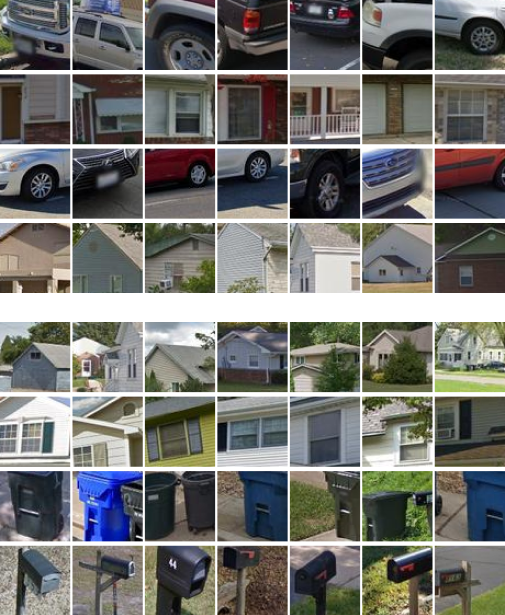}
        \caption{Clusters}
        \label{fig:finetuning-qualitative:b}
    \end{subfigure}
    \hfill
    \begin{subfigure}{0.26\linewidth}
        \centering
        \includegraphics[width=\textwidth, height=16em]{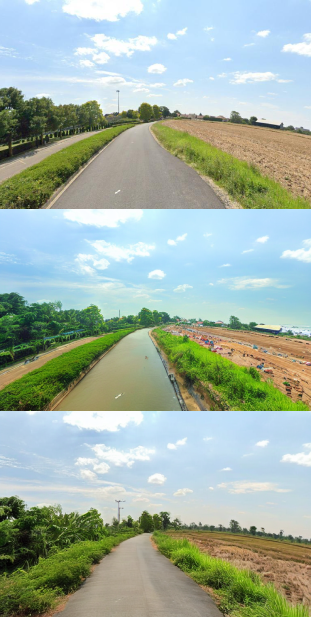}
        \caption{Parallel Translation}
        \label{fig:finetuning-qualitative:c}
    \end{subfigure}

    \label{fig:finetuning-qualitative}
    \caption{\textbf{Effect of finetuning}. \textbf{(a)} For the same USA image (top), finetuning changes the spatial allocation of typicality before (middle) and after (bottom) finetuning. \textbf{(b)} This results in different typical clusters (USA), which, after finetuning (bottom), select for more typical elements like mailboxes. \textbf{(c)} Translation (Sec.~\ref{sub:trends}) of a picture of a road from France (top) to Thailand without finetuning (middle) suffers from data biases in the base model turning the road into a river and erasing utility poles. After finetuning on the G\string^3 dataset (bottom), the translated image is more consistent with the original.
    }
    \vspace{\basefigem}
\end{figure}

\medskip
\noindent
\textbf{Effect of finetuning.}
Unsurprisingly, we found that finetuning the diffusion model on the dataset of interest was critical to the quality of our results.
First, on a given image, finetuning changes the spatial distribution of typicality, prioritizing elements more correlated with the training labels (see Fig.~\ref{fig:finetuning-qualitative:a}). Second, in Fig.~\ref{fig:finetuning-qualitative:b}, we show the most typical clusters identified before and after finetuning. The patches selected after finetuning avoid the biases in the training data of the base model and are more specific to the G~\string^3 dataset, identifying elements such as post-boxes. We also demonstrate this quantitatively in Section~\ref{disease} for our application to X-ray images. Third, finetuning enables better translation between labels (see Sec.~\ref{sub:trends}), as can be seen in Figure~\ref{fig:finetuning-qualitative:c}, allowing vegetation, roads, road tracks, and utility poles to be translated consistently across the class labels in the parallel dataset (which can be found in the supplementary material).


\begin{figure*}[t]
    \centering
    \begin{subfigure}{0.321\linewidth}
        \centering
        \includegraphics[width=\textwidth]{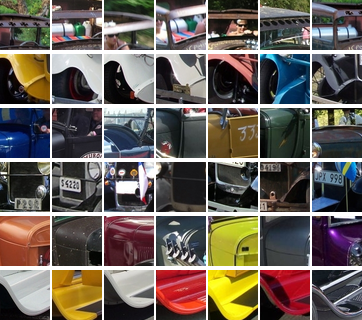}
        \caption{\textbf{\carcla}\textbf{s}}
        \label{cars:cluster:a}
    \end{subfigure}\hfill
    \begin{subfigure}{0.321\linewidth}
        \includegraphics[width=\textwidth]{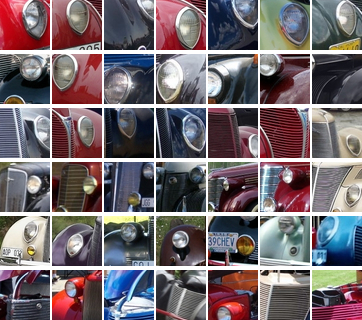}
        \caption{\textbf{\carclb}\textbf{s}}
        \label{cars:cluster:b}
    \end{subfigure}\hfill
    \begin{subfigure}{0.321\linewidth}
        \includegraphics[width=\textwidth]{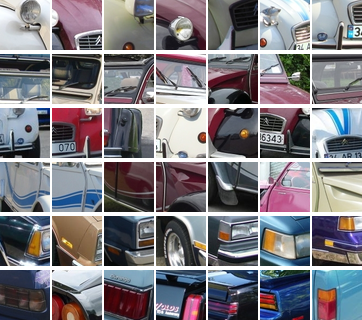}
        \caption{\textbf{\carcld}\textbf{s}}
        \label{cars:cluster:d}
    \end{subfigure}
    \caption{\textbf{Clusters of CarDB~\cite{cardb} visual elements.}
    Our visual summaries of typical car elements show elements unique to a period and elements that evolve with time. Evolving elements include the shapes of the car's body or headlights, which are parts of the 6 most {typical} clusters for most periods. More specific elements include running boards {in the 1920s} (\textbf{(a)}, {6th row}) or large engine side grills in the 1930s (\textbf{(b)}, {3rd, 4th and 6th row}). In the 1980s \textbf{(c)}, we observe two typical yet very discrete clusters of car design styles, of the curvy French 2CV (1-4 row) juxtaposed to the square American \textit{chevy}-style cars (5-6 rows).
    }
    \vspace{\basefigem}
    \label{fig:cars-clusters}
\end{figure*}

\begin{figure*}
    \centering
    \begin{subfigure}{0.321\linewidth}
        \includegraphics[width=\textwidth]{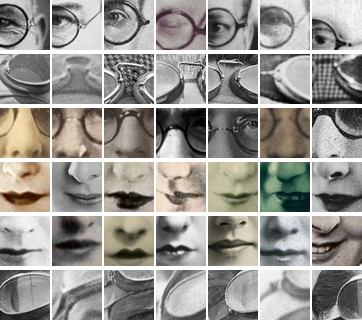}
        \caption{\textbf{\fttclb}\textbf{s}}
        \label{ftt:cluster:b}
    \end{subfigure}
    \hfill
    \begin{subfigure}{0.321\linewidth}
        \includegraphics[width=\textwidth]{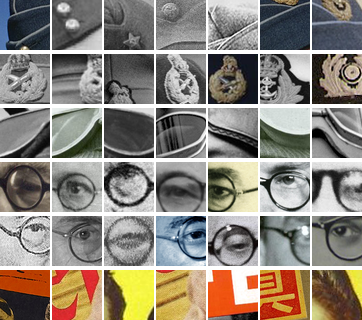}
        \caption{\textbf{\fttclc}\textbf{s}}
        \label{ftt:cluster:c}
    \end{subfigure}
    \hfill
    \begin{subfigure}{0.321\linewidth}
        \includegraphics[width=\textwidth]{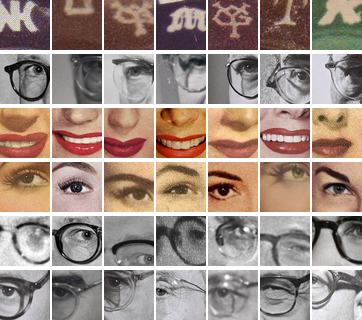}
        \caption{\textbf{\fttcld}\textbf{s}}
        \label{ftt:cluster:d}
    \end{subfigure}
    \caption{\textbf{Clusters of FTT~\cite{ftt} visual elements.}
Our cluster analysis of faces revealed that eyeglasses of varying designs are indicative of a portrait's decade throughout the history captured by FTT. Observing the 6 most typical clusters for the 1920s \textbf{(a)}, the 1940s \textbf{(b)}, and the 1950s \textbf{(c)}, we see how the shape of glasses is highly informative of each period. We also located fashion items that uniquely trended only in a particular period, such as aviator goggles in the 1920s (2nd row), military caps in the 1940s (1st and 2nd row), and baseball caps in the 1950s (1st row). Consistent with prior analysis~\cite{ginosar2017yearbooks}, we also found clusters corresponding to smiles and makeup.
    }
    \label{fig:ftt-clusters}
    \vspace{\basefigem}
\end{figure*}

\newcommand{\geowidth}{0.321}
\begin{figure*}
    \centering
    \begin{subfigure}{\geowidth\linewidth}
        \centering
        \includegraphics[trim={0 0 0px 0}, clip, width=\textwidth]{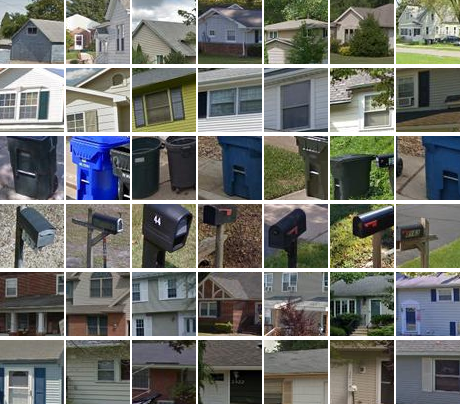}
        \caption{\textbf{\geocla}}
        \label{geo:cluster:a}
    \end{subfigure}
    \begin{subfigure}{\geowidth\linewidth}
        \includegraphics[trim={0 0 0px 0}, clip, width=\textwidth]{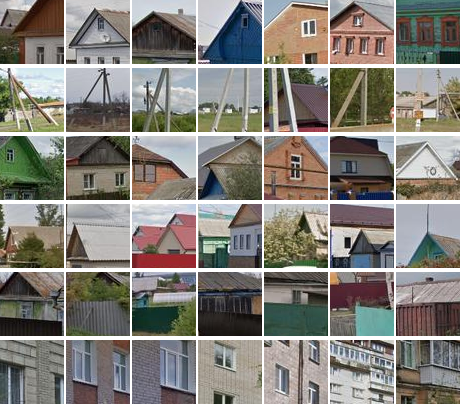}
        \caption{\textbf{\geoclb}}
        \label{geo:cluster:b}
    \end{subfigure}
    \begin{subfigure}{\geowidth\linewidth}
        \vspace{0.3em}
        \includegraphics[trim={0 0 0px 0}, clip, width=\textwidth]{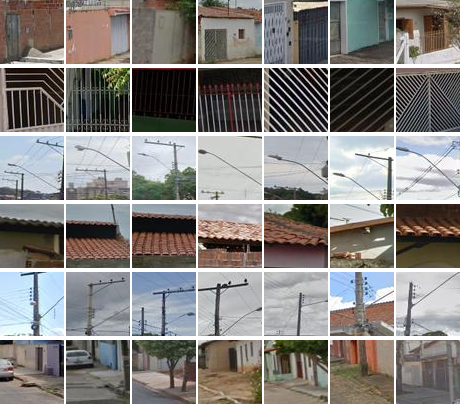}
        \caption{\textbf{\geoclc}}
        \label{geo:cluster:c}
    \end{subfigure}
    \begin{subfigure}{\geowidth\linewidth}
    	\vspace{0.5em}
        \includegraphics[trim={0 0 0px 0}, clip, width=\textwidth]{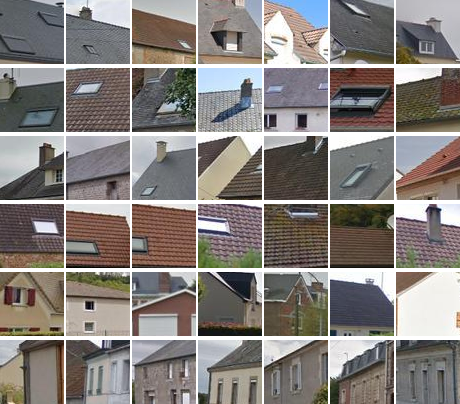}
        \caption{\textbf{\geocld}}
        \label{geo:cluster:d}
    \end{subfigure}
    \begin{subfigure}{\geowidth\linewidth}
        \includegraphics[trim={0 0 0px 0}, clip, width=\textwidth]{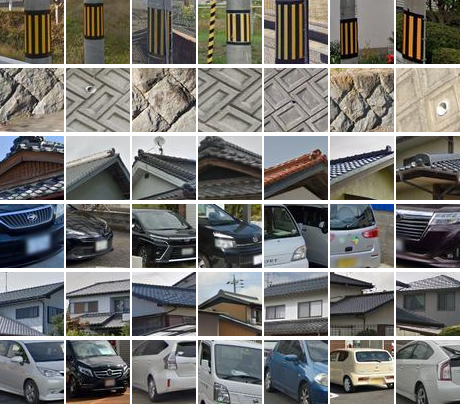}
        \caption{\textbf{\geocle}}
        \label{geo:cluster:e}
    \end{subfigure}
    \begin{subfigure}{\geowidth\linewidth}
        \includegraphics[trim={0 0 0px 0}, clip, width=\textwidth]{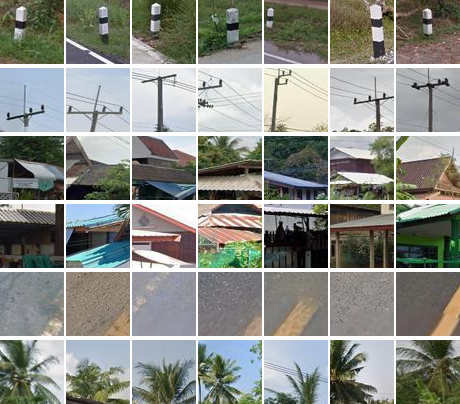}
        \caption{\textbf{\geoclf}}
        \label{geo:cluster:f}
    \end{subfigure}
    \caption{\textbf{Clusters of~G\string^3~ \cite{g3} visual elements.}
Our geographic clusters show a wide diversity of typical elements across different countries. We found architectural elements such as roofs, facades, or windows among the most typical elements in all countries. For example, \textbf{(a)} the ``double hung'' American windows (2nd row), \textbf{(d)} French roof windows (1st-4th row), or \textbf{(f)} covered pathways in Thailand (4th row). Utility poles are ranked second in Russia and Thailand and 5th in Brazil. We also found typical objects that are unique to a single country, such as \textbf{(a)} American garbage cans and post boxes (3rd, 4th row), \textbf{(c)} protective guard rails in Brazil (2nd row), \textbf{(e)} Japanese electricity warning signs and exterior wall tiles (1st, 2nd row), and \textbf{(f)} Thai Bollards (1st row). 
    }
    \label{fig:geo-clusters}
    \vspace{\basefigem}
\end{figure*}
\newcommand{\placeswidth}{0.321}
\begin{figure*}[t]
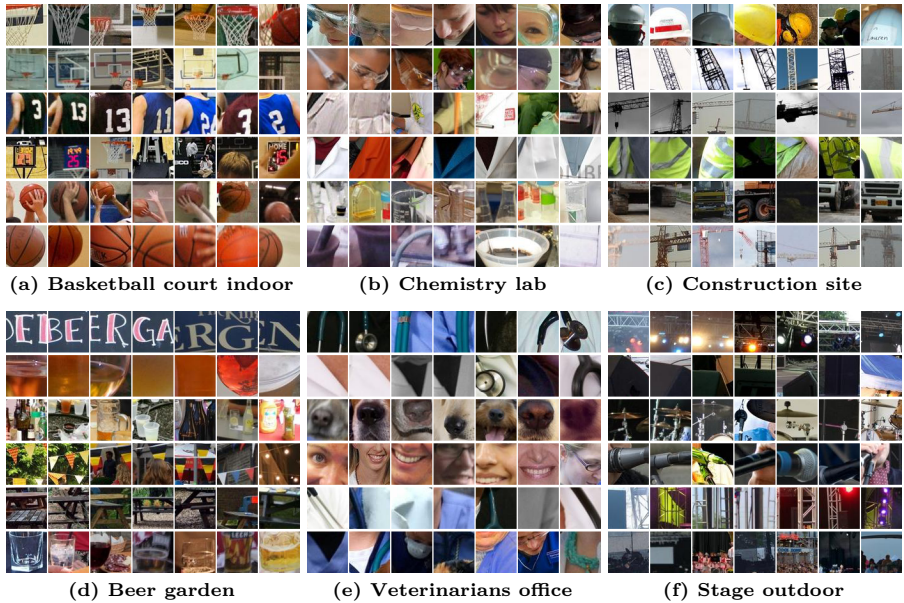

    \centering
    \begin{subfigure}{\placeswidth\linewidth}
        \centering
        \includegraphics[trim={0 0 0px 0}, clip, width=\textwidth]{images/places/\placesa__hard_limit_7__top_k_6__min_im_6_ranked.png}
        \caption{\textbf{\processstring{\placesa}}}
        \label{places:cluster:a}
    \end{subfigure}
    \begin{subfigure}{\geowidth\linewidth}
        \includegraphics[trim={0 0 0px 0}, clip, width=\textwidth]{images/places/\placesb__hard_limit_7__top_k_6__min_im_6_ranked.png}
        \caption{\textbf{\processstring{\placesb}}}
        \label{places:cluster:b}
    \end{subfigure}
    \begin{subfigure}{\geowidth\linewidth}
        \vspace{0.3em}
        \includegraphics[trim={0 0 0px 0}, clip, width=\textwidth]{images/places/\placesc__hard_limit_7__top_k_6__min_im_6_ranked.png}
        \caption{\textbf{\processstring{\placesc}}}
        \label{places:cluster:c}
    \end{subfigure}
    \begin{subfigure}{\geowidth\linewidth}
    	\vspace{0.5em}
        \includegraphics[trim={0 0 0px 0}, clip, width=\textwidth]{images/places/\placesd__hard_limit_7__top_k_6__min_im_6_ranked.png}
        \caption{\textbf{\processstring{\placesd}}}
        \label{places:cluster:d}
    \end{subfigure}
    \begin{subfigure}{\geowidth\linewidth}
        \includegraphics[trim={0 0 0px 0}, clip, width=\textwidth]{images/places/\placese__hard_limit_7__top_k_6__min_im_6_ranked.png}
        \caption{\textbf{\processstring{\placese}}}
        \label{places:cluster:e}
    \end{subfigure}
    \begin{subfigure}{\geowidth\linewidth}
        \includegraphics[trim={0 0 0px 0}, clip, width=\textwidth]{images/places/\placesf__hard_limit_7__top_k_6__min_im_6_ranked.png}
        \caption{\textbf{\processstring{\placesf}}}
        \label{places:cluster:f}
    \end{subfigure}
    \caption{\textbf{Clusters of~Places365 \cite{places365} visual elements.}
    Unlike the other datasets we analyze, each class label correlates with objects of different categories in the scenes dataset, as different scenes contain objects of different categories. Yet, our approach can still summarize a large variety of complex scenes with their unique typical elements. For example, in basketball courts \textbf{(a)}, our approach locates the basket (1st row), the backboard (2nd row), the jersey numbers (3rd row), the shot clock (4th row), a shoot (5th row), and the ball (6th row). Our approach can still focus and summarize the most critical elements even in more cluttered scenes like an outdoor stage, chemistry labs, or beer gardens. For example, in the case of outdoor stages \textbf{(f)}, we can see a lot of technical elements involved in their installation, including lights and top rails (1st row), monitor speakers (2nd row), microphones (4th row), and side rails (5th row).
    }
    \label{fig:places-clusters}
    \vspace{\basefigem}
\end{figure*}

\subsection{Clusters of Typical Visual Elements}
\label{sec:exp-clusters}

In this section, we analyze our visual summary of each dataset, obtained by clustering the typical visual elements for the different class labels. We show our summaries for Cars, Faces, Geo, and Scenes in Figs.~\ref{fig:cars-clusters},~\ref{fig:ftt-clusters},~\ref{fig:geo-clusters},~\ref{fig:places-clusters} respectively. For all datasets, we show the top-{6} typical elements of the top-{6} clusters ranked by the median typicality of their elements for selected class labels. We analyze the resulting clusters for each dataset inline in the figure captions for ease of viewing. Our complete clusters can be found in the supplementary material.

\medskip
\noindent
{\textbf{Comparison to Doersch \etal~\cite{paris2015}.} As the Matlab implementation of~\cite{paris2015} is obsolete and hardware-specific, we reimplement their method in Python and release this reimplementation with our code. In Fig.~\ref{fig:geod-clusters}, we show the results of this approach when applied to the same subset of the G\string^3 dataset as our approach. Similar to the original paper, we rank the trained detectors by \textit{discriminativeness}, \ie, the percentage of the top-50 final matches inside the positive set~\cite{paris2015}, and for each we show its top 6 matches. The results produced with the Doersch \etal method demonstrate more textures, appear much less semantic, and contain much more similar elements than ours. Note that the results in the original Doersch \etal paper do not show such failures, and in particular much less vegetation, simply because the paper used a curated and non-publicly available dataset of images focused on selected cities extracted from Google Street View.}
\newcommand{\geodwidth}{0.321}
\begin{figure*}[t]
    \centering
    \begin{subfigure}{\geodwidth\linewidth}
        \centering
        \includegraphics[trim={0 1766px 0px 0px}, clip, width=\textwidth]{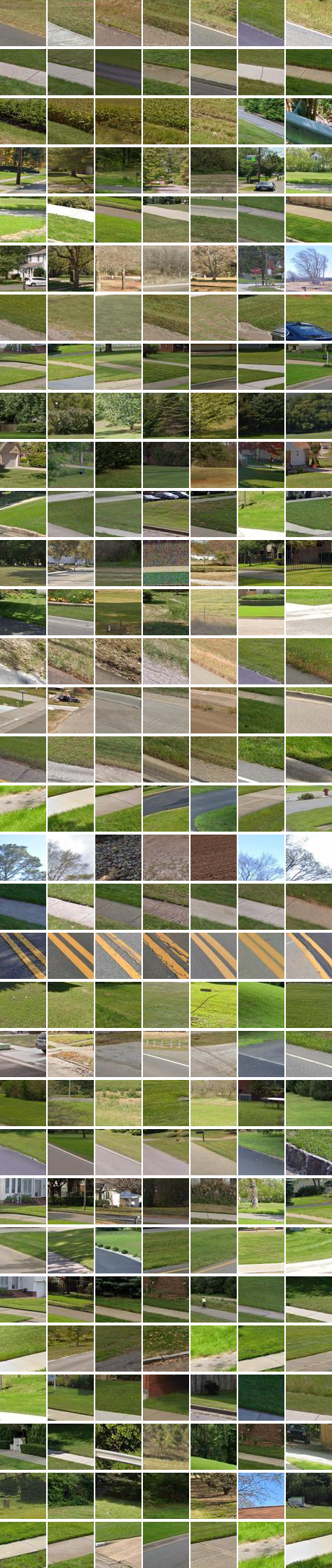}
        \caption{\textbf{\geocla}}
        \label{geod:cluster:a}
    \end{subfigure}
    \begin{subfigure}{\geodwidth\linewidth}
        \includegraphics[trim={0 1766px 0px 0px}, clip, width=\textwidth]{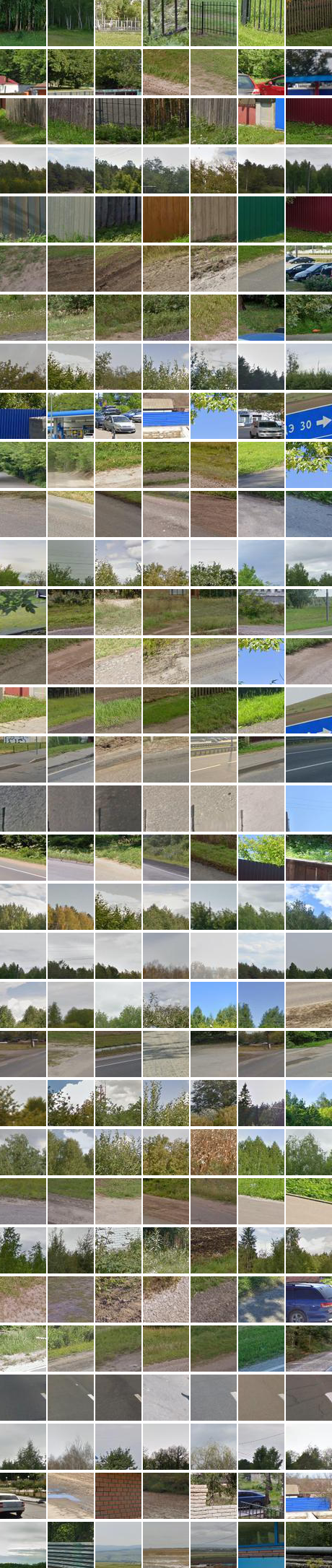}
        \caption{\textbf{\geoclb}}
        \label{geod:cluster:b}
    \end{subfigure}
    \begin{subfigure}{\geodwidth\linewidth}
        \vspace{0.3em}
        \includegraphics[trim={0 1766px 0px 0}, clip, width=\textwidth]{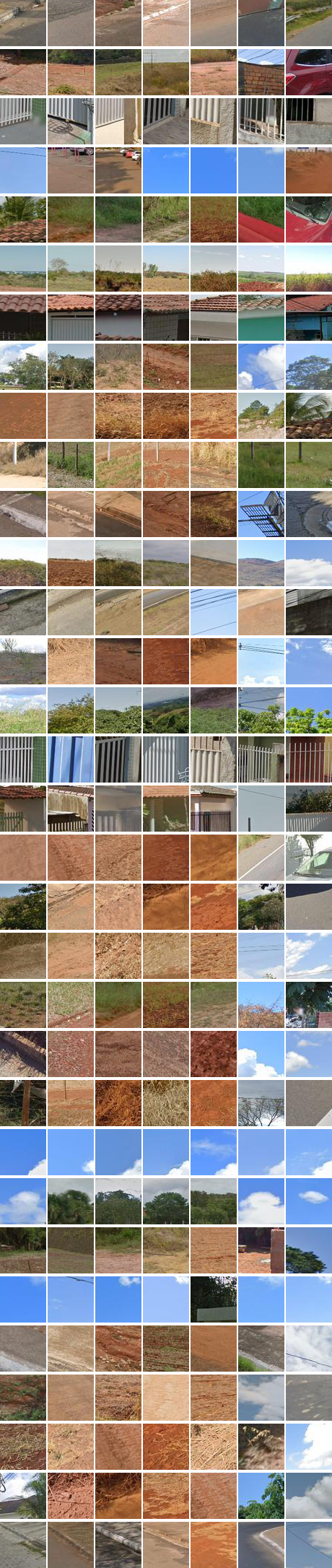}
        \caption{\textbf{\geoclc}}
        \label{geod:cluster:c}
    \end{subfigure}
    \begin{subfigure}{\geodwidth\linewidth}
    	\vspace{0.5em}
        \includegraphics[trim={0 1766px 0px 0}, clip, width=\textwidth]{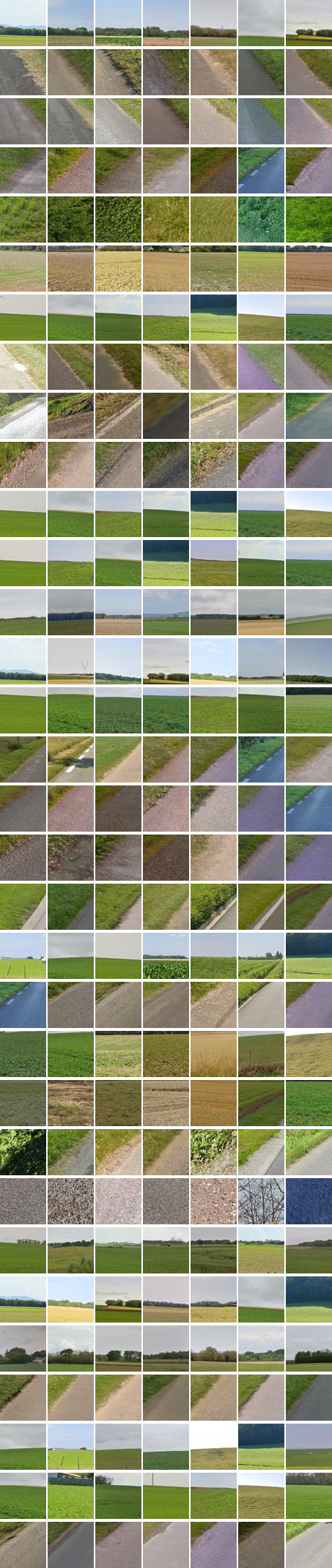}
        \caption{\textbf{\geocld}}
        \label{geod:cluster:d}
    \end{subfigure}
    \begin{subfigure}{\geodwidth\linewidth}
        \includegraphics[trim={0 1766px 0px 0}, clip, width=\textwidth]{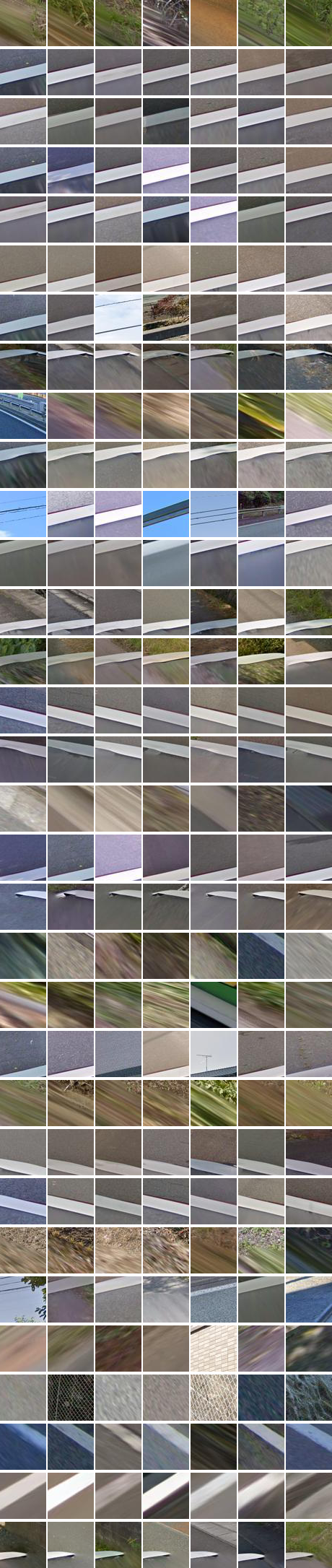}
        \caption{\textbf{\geocle}}
        \label{geod:cluster:e}
    \end{subfigure}
    \begin{subfigure}{\geodwidth\linewidth}
        \includegraphics[trim={0 1766px 0px 0}, clip, width=\textwidth]{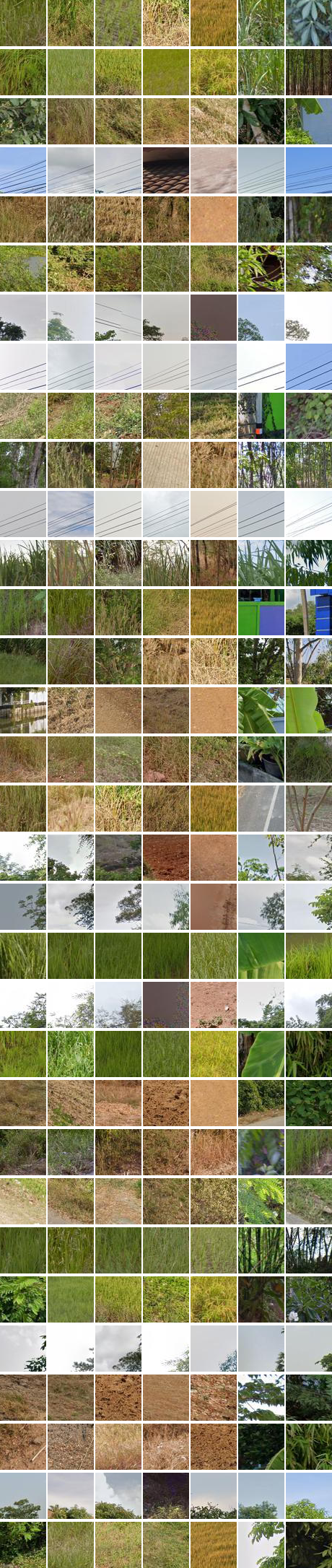}
        \caption{\textbf{\geoclf}}
        \label{geod:cluster:f}
    \end{subfigure}
    \caption{\textbf{Doersch~\etal, 2013~\cite{paris2015} results on ~G\string^3~\cite{g3}.} See text for details.}
    \label{fig:geod-clusters}
\end{figure*}



\subsection{Limitations}
\label{ss:lim}
Although our method makes the first step towards utilizing generative models for data mining, it comes with limitations. We visualize our two main failure modes in Fig.~\ref{fig:limitations}. First, clustering elements using k-means can lead to mixed clusters containing different categories of samples (Fig.~\ref{fig:limitations:a}) or produce repetitively similar clusters. Second, our method identified data artifacts (Fig.~\ref{fig:limitations:b}) that are related to noisy printing or scanning of old photographs or post-processing artifacts of street view images, which are highly typical but irrelevant to our purpose. Interestingly, in the case of street-view data such artifacts are suggested in GeoGuessr~\cite{geoguessr} advice websites~\cite{geodummy,geohints,plonkit}, as shortcuts for geolocation.
\begin{figure}
\centering
    \begin{subfigure}{0.46\linewidth}
        \includegraphics[width=\textwidth, height=2.5em]{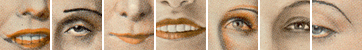}
        \includegraphics[width=\textwidth, height=2.5em]{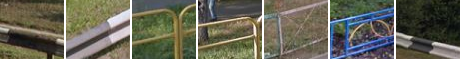}
        \caption{Mixed Clusters}
        \label{fig:limitations:a}
    \end{subfigure}
    \hspace{0.1em}
    \begin{subfigure}{0.46\linewidth}
        \includegraphics[width=\textwidth, height=2.5em]{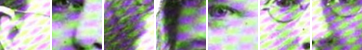}
        \includegraphics[width=\textwidth, height=2.5em]{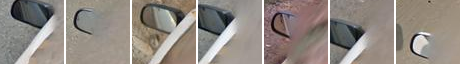}
        \caption{Mining Dataset Artifacts}
        \label{fig:limitations:b}
    \end{subfigure}
    \caption{\textbf{Limitations.} The two most common failure modes we observe are: \textbf{(a)} issues in clustering, for example, clusters that contain diverse visual content, or multiple clusters that correspond to the same concept; \textbf{(b)} typicality highlighting artifacts of the dataset. Discovering artifacts is an expected behavior and can be useful for some applications. 
    }
    \label{fig:limitations}
    \vspace{\basefigem}
\end{figure}

\section{Applications}
Our typicality score allows us to explore two different applications. First, in Section \ref{sub:trends}, we translate geographical elements across locations and mine typical translations. Then, in Section~\ref{disease}, we show how disease localization emerges from typicality when training to generate frontal X-rays of patients, of various diseases.

\subsection{Analyzing Trends of Visual Elements}
\label{sub:trends}
Having a diffusion model finetuned on a dataset of interest enables further applications that were not possible with previous visual mining approaches~\cite{paris2015, cardb, ftt, ginosar2017yearbooks}. One new application is the {summary of variation of typical visual elements} across different class labels. As a case study, we use the G\string^3 dataset to discover and summarize how \textit{co-typical} elements, such as windows, roofs, or license plates, vary across locations. We start by using our finetuned diffusion model to create a ``parallel dataset'', by translating all the images in our mining dataset to all  locations, then define a co-typicality measure.


\medskip
\noindent
\textbf{Generating a parallel dataset.}
We first use Plug and Play~\cite{tumanyan2023plug} to translate input images from one location to another, which we denote by $x^{c_{0}\rightarrow c}$, where $c_{0}$ is the initial country and $c$ is the target country. We translate 1000 images for each of the 10 selected countries to all others, resulting in 100K images, which we refer to as our parallel dataset. Performing translation using our finetuned model is critical for keeping scene elements consistent, as seen in Fig.~\ref{fig:finetuning-qualitative:c}.

We show in supplementary material how performing semantic segmentation for each image and its translations to different countries enables measuring statistical trends. For example, we can measure that translations to Thailand or Brazil add many potted plants, and translations to Nigeria add dirt roads and people. We can visually confirm those trends on our parallel dataset.

\medskip
\noindent
\textbf{Mining typical transformations across location.}
\label{cotypical} To further analyze our parallel dataset, we define a cross-location typicality measure to mine parallel translation of patches across locations. We define the co-typicality $\tbar$ as the median typicality across location:
\begin{equation}
\tbar(x) = \underset{c\in C}{\operatorname{med}}\bigl[\typicality(x^{c_{0}\rightarrow c}, c)\bigr],
\end{equation}
where $c_{0}$ is the true label of the patch $x$ and the median is computed over all {countries} in our set of 10 analyzed countries, $C$.

\def\wfl{1.2em}

\begin{figure*}[t]
    \includegraphics[width=1.0\textwidth]{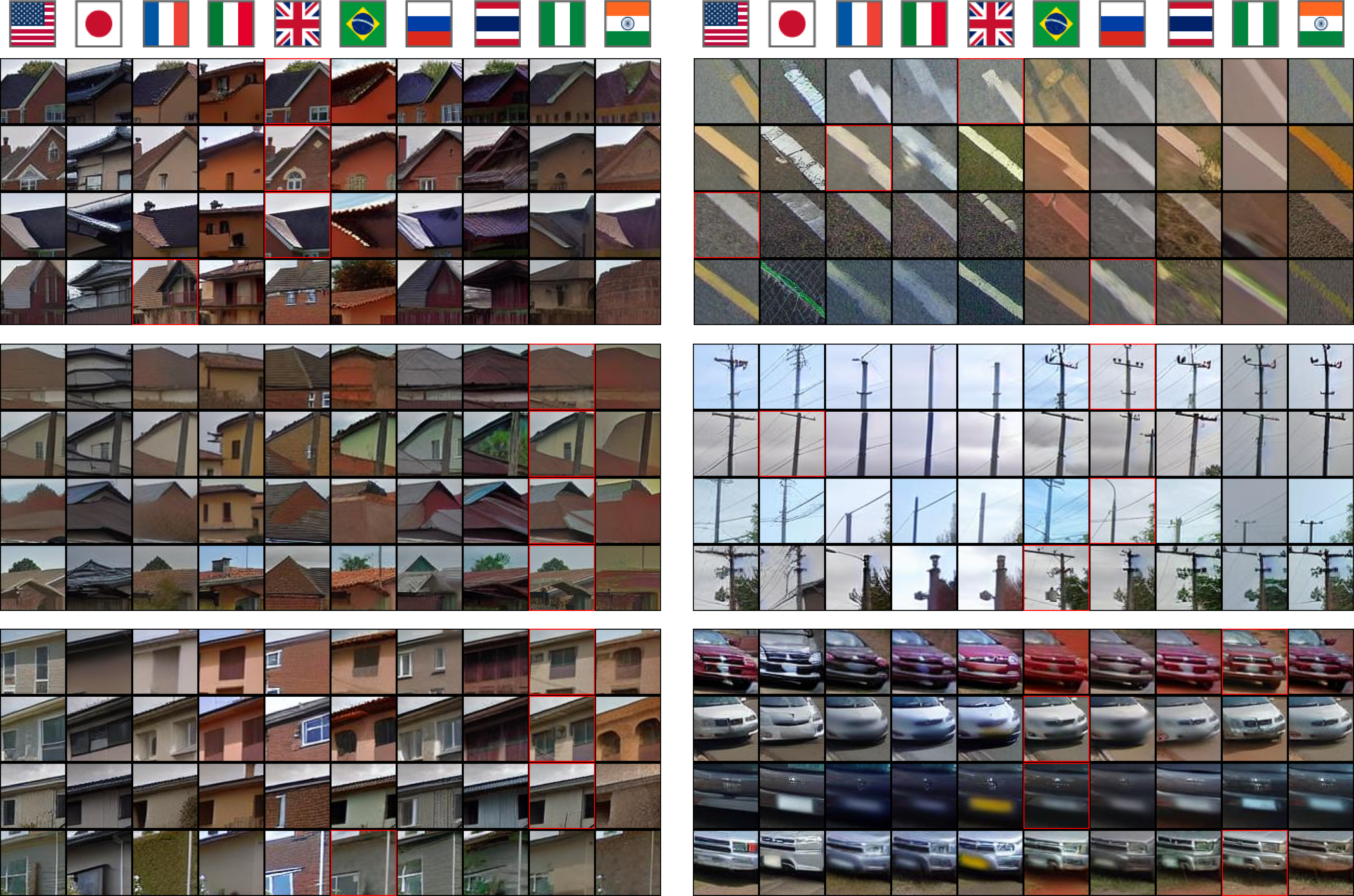}
   \caption{\textbf{{Clustering typical translations of elements across countries}.} Ranking translated visual elements according to $\tbar$ and clustering the translated sequences results in groups of elements with similar variations. We show elements from 6 selected clusters out of 32. The {source} image for each sequence is highlighted in \textcolor{red}{red}. See text for details.}
    \vspace{\basefigem}
    \label{coclus}
\end{figure*}

We can now ask: What visual elements are typical of a certain place and whose translation remains typical of another location? Instead of ranking single patches, we now rank a whole sequence of $|C|$ patches translated across locations according to $\tbar$. We represent this sequence by concatenating the DIFT features of each patch~\cite{dift}. {To facilitate clustering, we first project the DIFT features of each patch from 1280 dimensions to 32 using UMAP \cite{umap}}. To keep the same proportion of typical patches with respect to the number of analyzed images/sequences as in Section~\ref{sec:exp-clusters}, we cluster the 10,000 visual elements with the highest co-typicality.

We display our results in Fig.~\ref{coclus}, where for 6 selected clusters, we show in rows the four translated sequences closest to the cluster mean, highlighting in red the original image in each sequence. On the left column of Fig.~\ref{coclus}, we show changes in typical architectural elements, such as gables, roofs, and windows. In contrast, on the right we show regulation-related elements, such as road tracks, utility poles, and license plates. Our approach allows us to both locate and visualize how common visual elements would vary from place to place, even though an exact match does not exist in the original data. For example, roofs typically turn dark brown when translated to the UK and black when translated to Japan.

\subsection{Analysis of Medical Images}
\label{disease}
\begin{figure}[t]
    \centering
    \newlength{\colimagewidth}
    \setlength{\colimagewidth}{0.13\textwidth}
    
    \newlength{\rowimageheight}
    \setlength{\rowimageheight}{3\colimagewidth} 
    
    \newcommand{\colheader}[1]{
        {\tiny \centering #1}
    }
    
    \newcommand{\rowlabel}[1]{
        {\raisebox{0.1\rowimageheight}{#1}}
    }

    \begin{tabular}{cccccccc}
         & \colheader{Mass} & \colheader{Cardiomegaly} & \colheader{Nodule} & \colheader{Effusion} & \colheader{Atelectasis} & \colheader{Pneumonia} & \colheader{Pneumothorax}\\
        \makebox[7pt]{\hspace{-28pt}\raisebox{105pt}{gt.}} \hspace{-26pt}\raisebox{65pt}{pt.} \hspace{-14pt}\raisebox{20pt}{ft.}
        & \includegraphics[width=\colimagewidth, height=\rowimageheight]{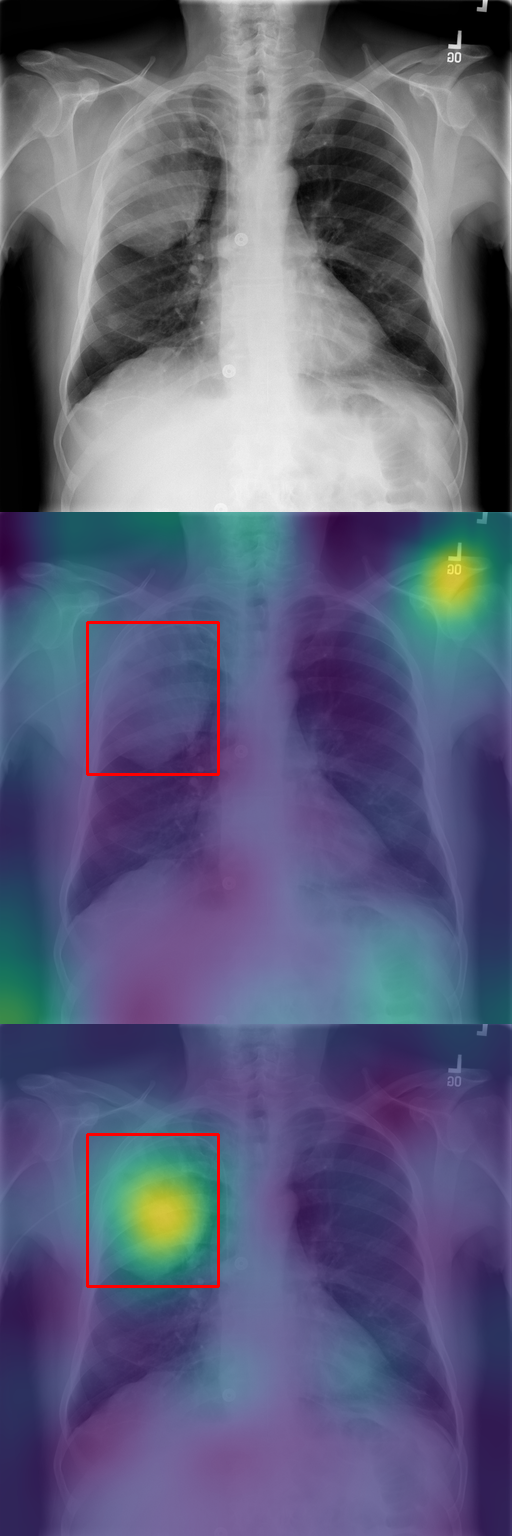}
        & \includegraphics[width=\colimagewidth, height=\rowimageheight]{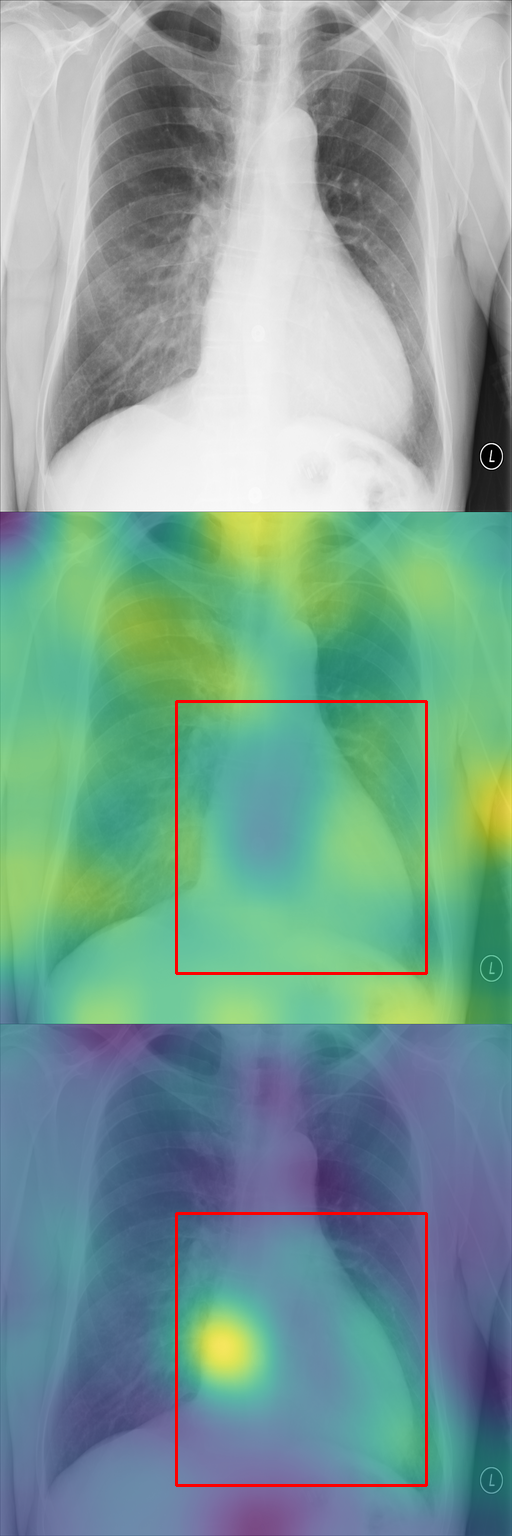}
        & \includegraphics[width=\colimagewidth, height=\rowimageheight]{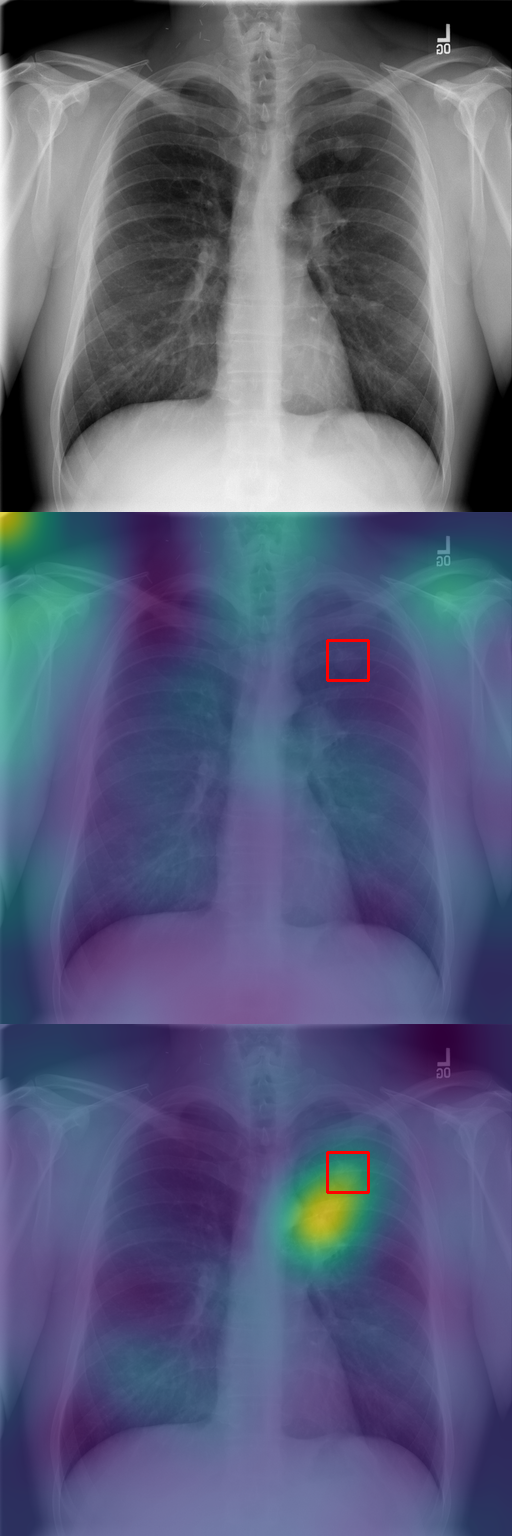}  
        & \includegraphics[width=\colimagewidth, height=\rowimageheight]{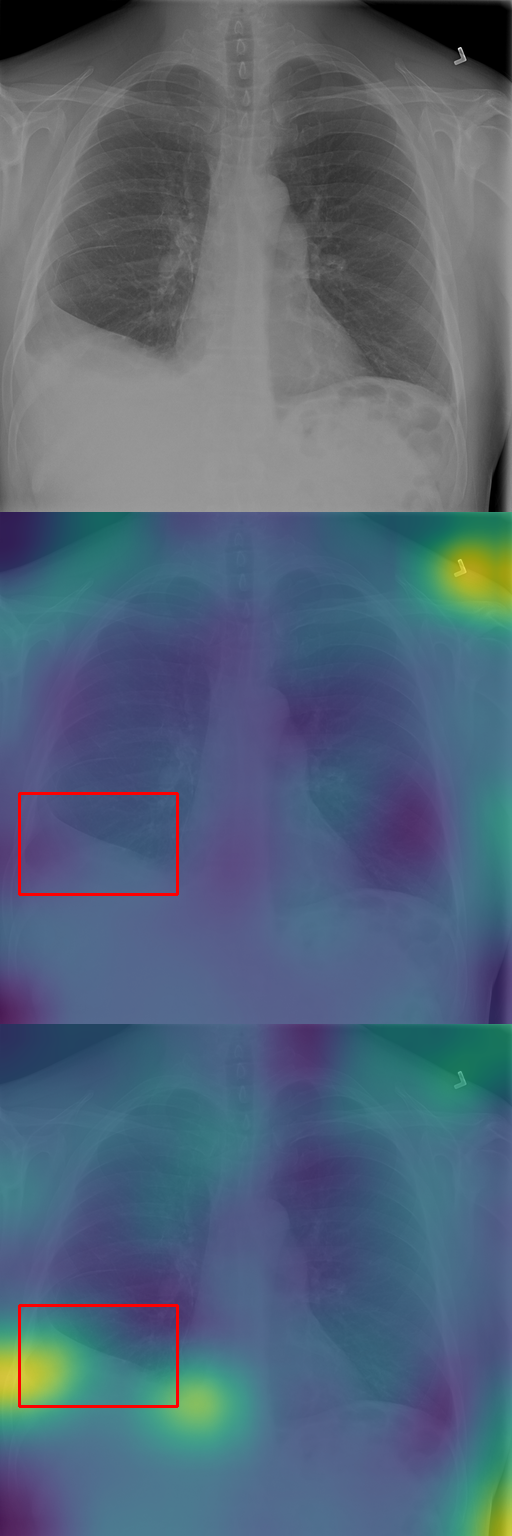} 
        & \includegraphics[width=\colimagewidth, height=\rowimageheight]{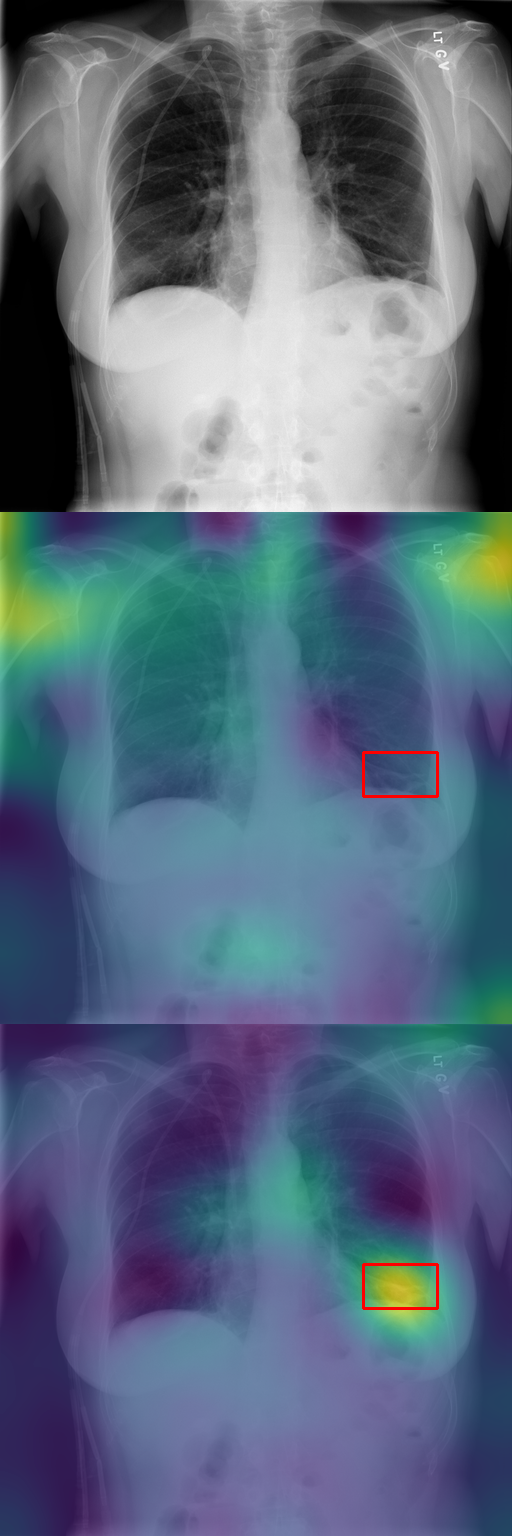}
        & \includegraphics[width=\colimagewidth, height=\rowimageheight]{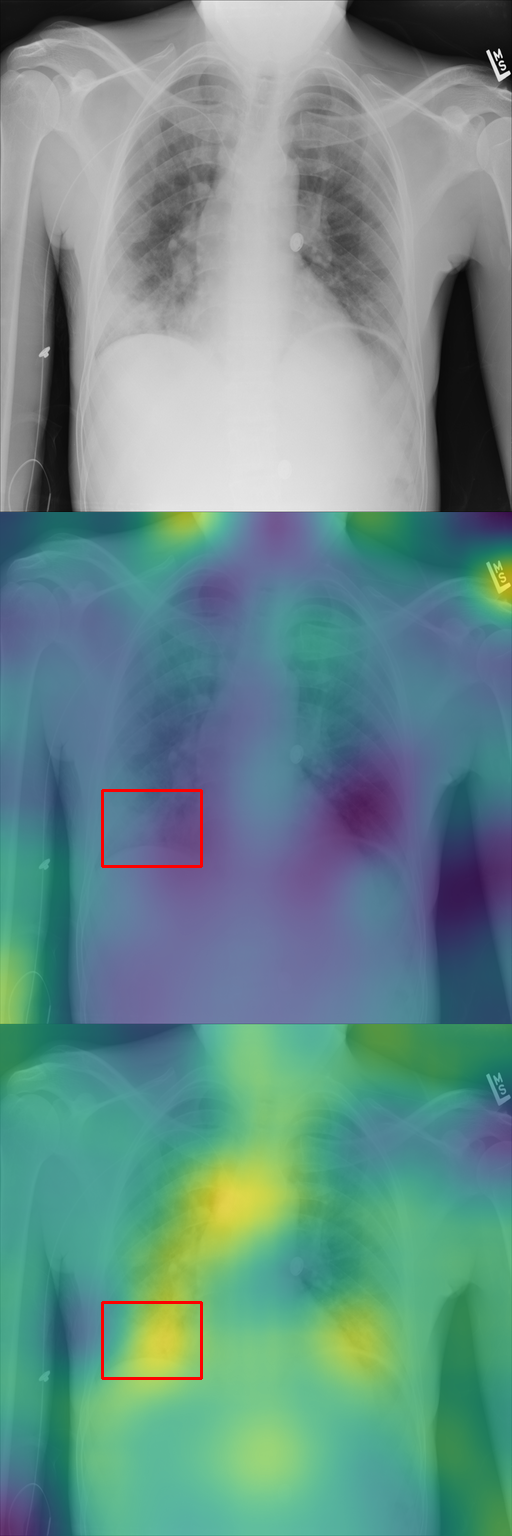} 
        & \includegraphics[width=\colimagewidth, height=\rowimageheight]{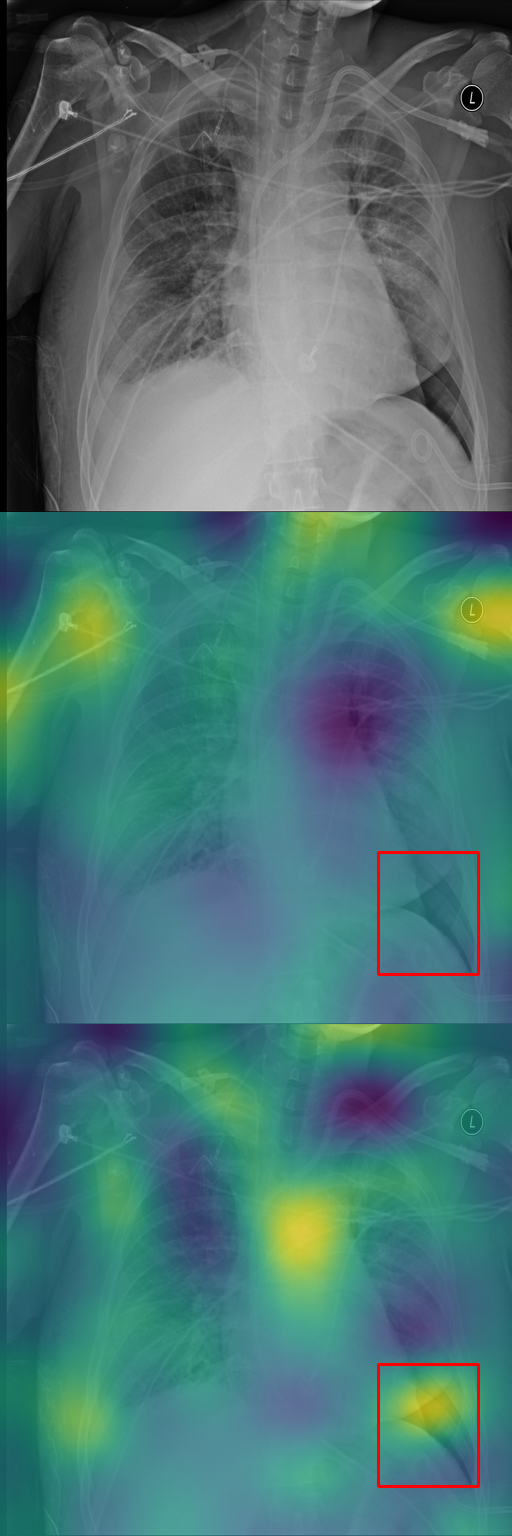}\\\cdashline{2-8}
        & \colheader{2\% $\uparrow$ 16.6\%} & \colheader{6\% $\uparrow$ 16.2\%} & \colheader{0\% $\uparrow$ 8.2\%} &  \colheader{3.3\% $\uparrow$ 7.5\%} & \colheader{1.3\% $\uparrow$ 6.3\%} &  
        \colheader{4\% $\uparrow$ 7.5\%} & \colheader{3.\% $\uparrow$ 6.5\%}\\
        
    \end{tabular}



    \caption{\textbf{Localizing abnormal areas in medical images.} We visualize typicality when finetuning our model on the CXR8 dataset of thorax diseases \cite{wang2017chestx}. After fine-tuning (ft.), we can see a clear focus of the typicality score on expert annotated areas (\textcolor{red}{red} boxes) for each disease, while initial predictions from {the pretrained Stable Diffusion V1.5} model (pt.) are mostly noise. {Images are ordered by AUC-PR after finetuning \cite{aucpr}. With $\uparrow$ we delimitate performance before and after finetuning, in the last row.}
    }
    \label{fig:finetuning-quantitative}    
    \vspace{\basefigem}
\end{figure}

{In Section \ref{ss:typicality-eval}, we discussed how typicality helps find relevant patches for an input label. In this section, we test this idea on completely different images: X-rays of patients who may suffer from a combination of various} thorax diseases. We finetune Stable Diffusion on the ChestX-ray8 dataset~\cite{wang2017chestx} containing 108,948 frontal-view X-ray images annotated with 14 single-word disease-name labels. Experts annotated a test set of 879 images with 7 diseases with rectangular regions of interest (ROI) for each disease. For each image, we compute typicality per latent feature, interpolate the resulting typicality to the input dimension, and blur the resulting typicality map for visualization.  In Fig.~\ref{fig:finetuning-quantitative}, we show the resulting typicality maps together with the ROI annotation before and after finetuning. Finetuning clearly improves the localization. We quantify this effect by computing the area under the precision recall-curve \cite{aucpr} (AUC-PR) associated with the ROIs. As reported in Fig.~\ref{fig:finetuning-quantitative}, we see consistent improvement of this measure when finetuning the network (from 3.2\% to 9.6\%), ranging from +3.5\% for Pneumonothorax (from 3\% to 6\%) to +14.6\% for Mass (from 2\% to 16.6\%), which are respectively the least and most localized diseases. Similar to our other experiments, finetuning uses only image labels without localization supervision.




\section{Conclusion}
We presented a novel use of diffusion models as visual mining tools. We defined a typicality measure using a pretrained stable diffusion model finetuned for conditional image synthesis. We used typicality to mine visual summaries of four datasets, tagged by year or location. We further showed that we can use our typicality measure to localize abnormalities in medical data and extend it to discover trends in the variations of translated visual elements within a generated parallel dataset. In summary, our work presents a novel approach to visual data mining, enabling scaling to datasets significantly more extensive and diverse than those showcased in prior works, as demonstrated by our experiments. 

\medskip
\medskip
\noindent \textbf{Acknowledgments.\;}  This work was partially supported by the European Research Council (ERC project DISCOVER, number 101076028) and leveraged the HPC resources of IDRIS under the allocation AD011012905R1, AD0110129052 made by GENCI. It was also partially supported by ONR MURI. We thank Grace Luo for data, code, and discussion; Loic Landreu and David Picard for insights on geographical representations and diffusion; Carl Doersch, for project advice and implementation insights; Sophia Koepke for feedback on our manuscript. 
\pagebreak\pagebreak

\bibliographystyle{splncs04}
\bibliography{main}
\end{document}